\documentclass[10pt,twocolumn,letterpaper]{article}

\usepackage{iccv}
\usepackage{times}
\usepackage{epsfig}
\usepackage{graphicx}
\usepackage{amsmath}
\usepackage{amssymb}
\usepackage{multirow}
\usepackage{adjustbox}
\usepackage{makecell}
\usepackage{algorithm} 
\usepackage{algpseudocode} 
\usepackage{xcolor}
\usepackage{color, colortbl}
\usepackage{booktabs}
\usepackage{tablefootnote}
\usepackage{kotex} 
\definecolor{LightGray}{gray}{0.90}

\usepackage{bm}

\usepackage[breaklinks=true,bookmarks=false]{hyperref}

\iccvfinalcopy 

\def\iccvPaperID{****} 

\ificcvfinal\fi

\begin{document}

\title{Winning the ICCV'2021 VALUE Challenge: \\Task-aware Ensemble and Transfer Learning with Visual Concepts}

\author{Minchul Shin$^1$ Jonghwan Mun$^1$ Kyoung-Woon On$^1$\textsuperscript{$\dagger$} Woo-Young Kang$^1$\textsuperscript{$\dagger$} Gunsoo Han$^1$\textsuperscript{$\dagger$} Eun-Sol Kim$^2$\thanks{This work was done while at Kakao Brain}\\
$^1$Kakao Brain \quad \quad \quad $^2$Hanyang University\\
{\tt\small \{craig.starr,jason.mun,kloud.ohn,edwin.kang,coco.han\}@kakaobrain.com, eunsolkim@hanyang.ac.kr}
}

\maketitle
\begingroup\renewcommand\thefootnote{$\dagger$}
\footnotetext{Equal contribution}
\endgroup
\ificcvfinal\thispagestyle{empty}\fi

\begin{abstract}
The VALUE (Video-And-Language Understanding Evaluation) benchmark is newly introduced to evaluate and analyze multi-modal representation learning algorithms on three video-and-language tasks: Retrieval, QA, and Captioning.
The main objective of the VALUE challenge is to train a task-agnostic model that is simultaneously applicable for various tasks with different characteristics.
This technical report describes our winning strategies for the VALUE challenge: 1) single model optimization, 2) transfer learning with visual concepts, and 3) task-aware ensemble. The first and third strategies are designed to address heterogeneous characteristics of each task, and the second one is to leverage rich and fine-grained visual information. We provide a detailed and comprehensive analysis with extensive experimental results. Based on our approach, we ranked first place on the VALUE and QA phases for the competition.
\end{abstract}

\section{Introduction}
In recent years, one of the major research streams is pre-training of a foundation model (\eg, BERT~\cite{BERT}, GPT-3~\cite{GPT3}, and CLIP~\cite{CLIP}) followed by transfer learning to multiple downstream tasks.
Following that, pre-training multi-modal representation (\eg, MIL-NCE~\cite{mil-nce} and HERO~\cite{HERO}) for videos is also widely studied using large-scale datasets.
However, due to the lack of a benchmark, algorithms of pre-training multi-modal representation for videos rely on different downstream tasks, making the comparison of algorithms difficult.

\begin{figure}[t]
    \begin{center}
    \scalebox{0.93}{
	    \includegraphics[width=\linewidth]{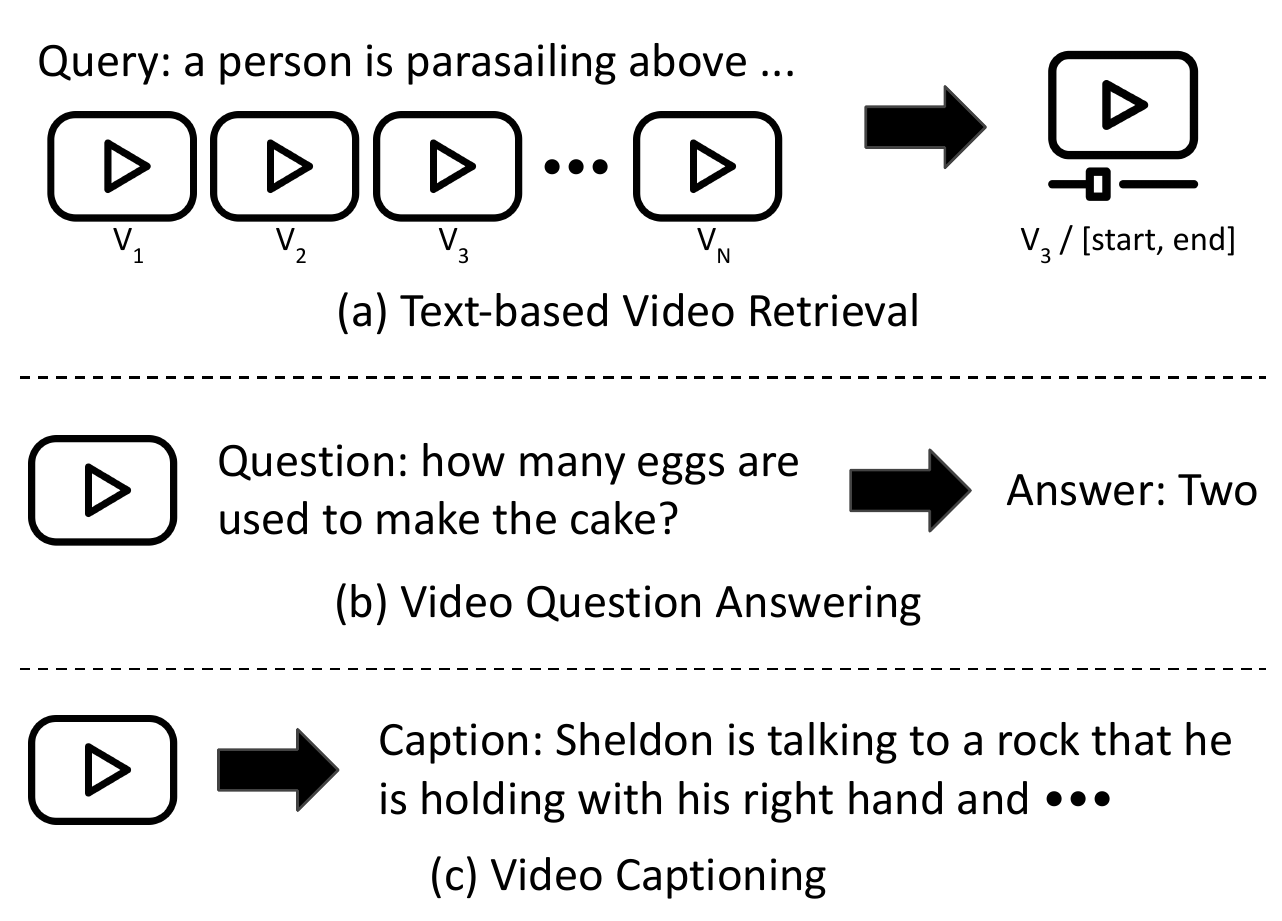}
	}
	\end{center}
	\vspace{-0.3cm}
	\caption{
	Illustration of three video-and-language tasks: (a) text-based video retrieval, (b) video question answering, and (c) video captioning.
	}
	\label{fig:task_description}
\end{figure}

Motivated by this, the VALUE benchmark (or challenge) is proposed to measure the generalization ability and versatility of a multi-modal pre-trained model for videos.
There are two critical characteristics of the VALUE benchmark compared to other benchmarks. First, it comprehensively measures the generalization ability of models over the popular video and language understanding tasks, including video retrieval, video question answering, and video captioning, as presented in Figure~\ref{fig:task_description}. 
These three macro-tasks are defined by eleven tasks based on 6 widely used datasets (TV~\cite{tvqa}, HowTo100M~\cite{howto100m}, YouCook2~\cite{youcook2}, VATEX~\cite{vatex}, VIOLIN~\cite{violin}, and VLEP~\cite{vlep}) covering a broad range of video genres, lengths and data volume. 
Second, videos addressed in the VALUE benchmark have multi-modal video inputs, including frames, audio, and textual information. 
In contrast, most of the existing works tend to focus only on visual cues.
Therefore, the VALUE benchmark deals with multi-modal multi-task video data.

\begin{figure*}[!t]
    \begin{center}
        \scalebox{0.9}{
    	    \includegraphics[width=\linewidth]{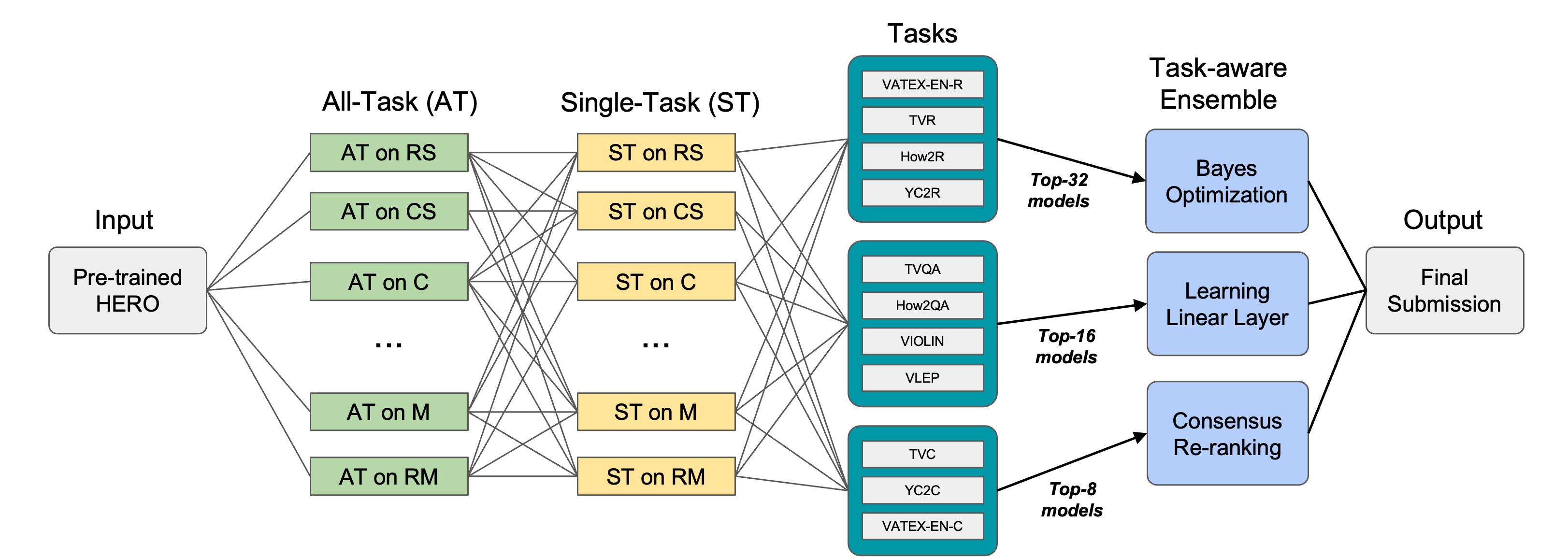}
        }
	\end{center}
	\caption{The entire pipeline of our approach. The model is initialized with the HERO model pre-trained using TV~\cite{tvqa} and HowTo100M~\cite{howto100m} datasets. 
	When the selected fine-tuning strategy is single-task (ST) only, the process of all-task (AT) fine-tuning is skipped.
	}
	\label{fig:main_diagram}
\end{figure*}

In this report, we describe our three winning strategies for the VALUE challenge. The first strategy is single model optimization. Since each task has different behavior, it would be sub-optimal to use the same training configurations for the individual task. Through extensive experiments, we found the best-performing training configurations for each task, and that was to identify the best combination of visual features and fine-tuning strategies.
In addition, we exploit visual concepts (\textit{objects} and \textit{associated attributes}) as an auxiliary source and combine them with the global clip-level visual representations.
It allows our model to leverage rich and fine-grained visual information.
Finally, we adopt the task-aware ensemble strategy. Since the output format of a model varies depending on the task, we apply different ensemble strategies that are specialized for each task. 
Through these three strategies, we ranked first place on the VALUE and QA phases.

\begin{table*}[h]
    \caption{Optimized fine-tuning hyperparameters of HERO for individual tasks.}
    \vspace{+0.12cm}
    \centering
    \begin{adjustbox}{width=0.98\textwidth}
        \begin{tabular}{l|cccc|cccc|ccc}
            \toprule
            \multirow{2}{*}{Hyperparameter} & \multicolumn{4}{c|}{Retrieval} & \multicolumn{4}{c|}{QA} & \multicolumn{3}{c}{Captioning} \\
            & TVR & How2R & YC2R & VATEX-EN-R & TVQA & How2QA & VIOLIN & VLEP & TVC & YC2C & VATEX-EN-C \\
            \hline \hline
            Learning rate & 1e-04 & 1e-04 & 7e-05 & 7e-05 & 5e-05 & 5e-05 & 3e-05 & 5e-05 & 1e-04 & 1e-04 & 1e-4 \\
            \# GPUs & 4 & 4 & 4 & 2 & 8 & 4 & 2 & 4 & 2 & 2 & 2 \\
            Batch size per GPU & 16 & 16 & 24 & 256 & 2 & 4 & 16 & 4 & 8 & 32 & 64 \\
            Gradient accumulation steps & 8 & 8 & 4 & 2 & 32 & 4 & 2 & 4 & 2 & 2 & 8 \\
            Effective batch size & 512 & 512 & 384 & 1024 & 512 & 64 & 64 & 64 & 32 & 128 & 1024 \\
            \# fine-tuning steps & 5,000 & 3,000 & 4,000 & 4,000 & 10,000 & 2,000 & 6,000 & 1,000 & 7,000 & 7,000 & 7,000 \\
            \bottomrule
        \end{tabular}
    \end{adjustbox}
    \label{tab:training_config}
\end{table*}

\section{Our Approach}
In this section, we introduce our winning solution in detail.
Our base model is HERO~\cite{HERO}, and we begin with the starter code\footnote{https://github.com/VALUE-Leaderboard/StarterCode} provided by the competition organizers. Based on this, we focus on the fine-tuning stage for individual tasks and improve the performance through the following three strategies: 1) single model optimization, 2) transfer learning with visual concepts, and 3) task-aware ensemble. 
We opt for the best model based on the validation score, assuming that the distribution of validation and test sets are similar.
We illustrate the entire process of our pipeline in Figure~\ref{fig:main_diagram}.

\subsection{Revisiting VALUE Challenge}

The VALUE benchmark is a collection of video-and-language datasets on multi-channel videos (\eg, video and subtitle) across various video domains and genres.
The benchmark contains 11 tasks (TVR, How2R, YC2R, VATEX-EN-R, TVQA, How2QA, VIOLIN, VLEP, TVC, YC2C, and VATEX-EN-C) each of which belongs to one of 3 video-and-language macro-tasks---(a) text-based video retrieval (Retrieval), (b) video question answering (QA) and (c) video captioning (Captioning)---as illustrated in Figure~\ref{fig:task_description}.
In this competition, the raw videos are not provided due to the license issue. Instead, the competition organizers provided raw text data (\ie, subtitles) and eight different types of clip-level visual features (ResNet+SlowFast, ResNet+MIL-NCE, Clip-ViT+SlowFast, Clip-ViT+MIL-NCE, ResNet, SlowFast, Clip-ViT, MIL-NCE) extracted from different pre-trained models (ResNet~\cite{resnet}, SlowFast~\cite{slowfast}, Clip-ViT~\cite{clip-vit}, MIL-NCE~\cite{mil-nce}).
Note that the plus (+) mark indicates the concatenation for multiple features.

Under the constraints, we choose the HERO model as our baseline due to the following two reasons:
First, the starter code based on HERO provides detailed training configurations for each task.
Second, we encounter enormous data loss when downloading raw videos from YouTube due to various reasons (\eg, broken URLs, blocked regions, etc.);
it makes the pre-training of any other algorithm from raw videos intractable.
HERO follows the two-stage training: (1) self-supervised learning on large-scale data, and (2) supervised fine-tuning stage on relatively smaller-scale data.
Given limited resources and time, we set our objective to improve the fine-tuning stage because pre-training a single HERO model takes three weeks, even with 16 GPUs as the original paper stated.
However, we believe that pre-training of HERO could be more important and influential than fine-tuning for achieving a higher score in the competition.
We refer interested readers to \cite{HERO} for more details about the structure of HERO and the way to apply it for different tasks.

\subsection{Single Model Optimization}
\label{sec:single_model_optimization}

Assuming our model is already pre-trained, we start by optimizing training configurations to fit each of the target tasks best;
the training hyper-parameters for individual tasks are summarized in Table~\ref{tab:training_config}.
In addition, we search for the best combination of visual features and fine-tuning strategies, which will be described below.

\paragraph{Visual feature adaptation.}
Note that as discovered by Li~\etal~\cite{value}, although HERO is trained using ResNet+SlowFast visual feature, fine-tuning the model with different types of visual features could improve the performance.
Therefore, choosing the visual features working best for the target task becomes vital for the higher score during the fine-tuning stage.
Therefore, we perform extensive experiments and identify the best-adapted feature out of eight visual features for individual tasks.

\paragraph{Fine-tuning strategies.}
According to the baseline paper of VALUE~\cite{value}, the authors explored two different ways of fine-tuning: \textit{ST} and \textit{AT}$\rightarrow$\textit{ST}. 
Specifically, ST (single-task) means that the HERO model is fine-tuned with the target task only, whereas AT$\rightarrow$ST (all-task to single-task) performs multi-task learning over all eleven tasks followed by further fine-tuning for a single target task.
We found that the best strategy is not uniform but rather different for each target task, as we describe in Table~\ref{tab:single_ensemble_model}.
Based on these observations, we exhaustively search for the best combinations for each task.
The best-performing combinations for each task are shown in Table~\ref{tab:single_ensemble_model}.

\subsection{Transfer Learning with Visual Concepts}
\label{sec:transfer_learning_vcept}
\begin{figure}[t]
    \begin{center}
    \scalebox{0.88}{
	    \includegraphics[width=\linewidth]{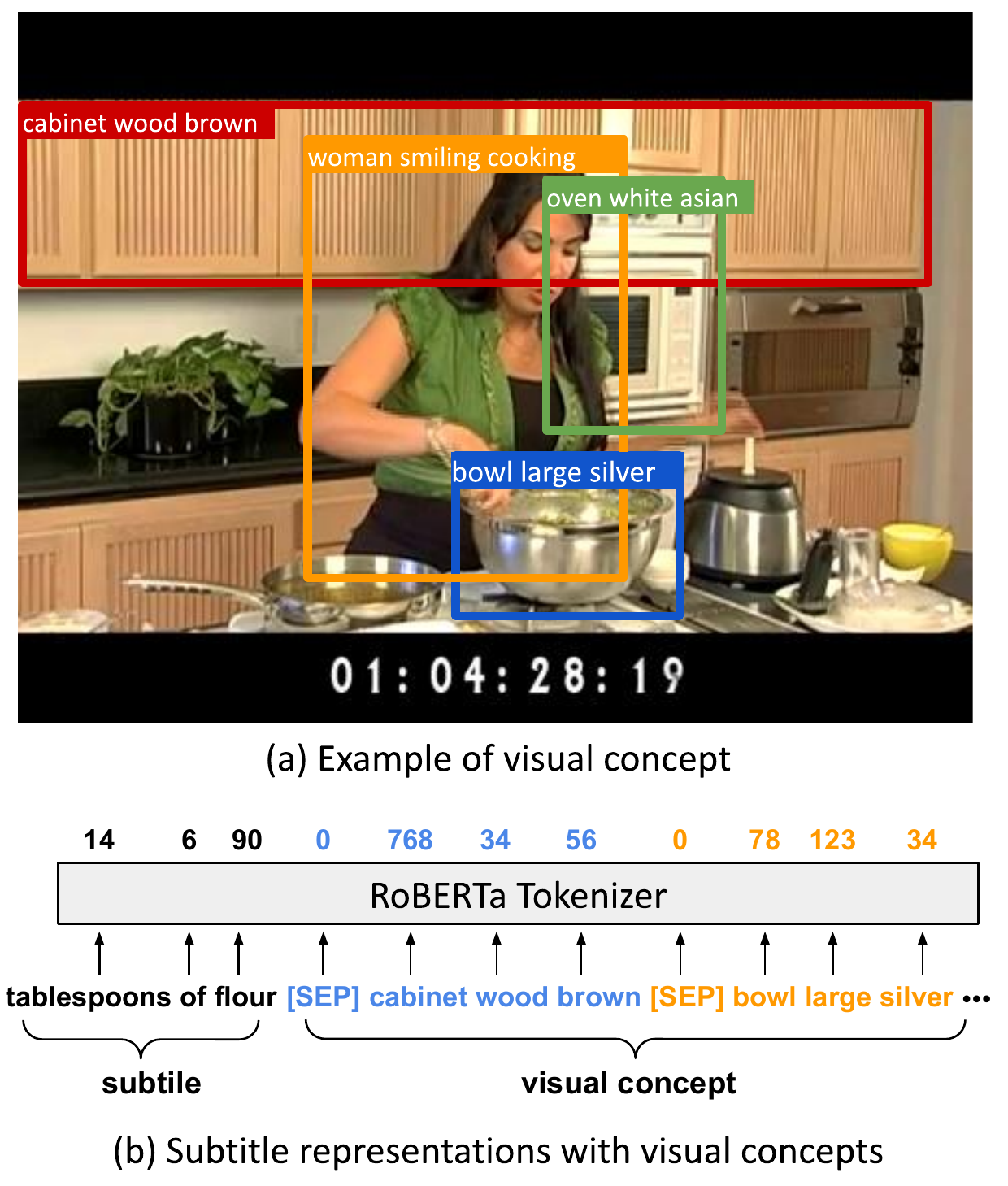}
	}
	\end{center}
	\vspace{-0.2cm}
	\caption{
	Example of visual concepts extracted from a frame in the YouCook2 video.
	The extracted visual concepts are exploited as additional language sources by attached with subtitles.
	}
	\label{fig:vcept}
\end{figure}

Visual concepts are often used for many vision-and-language tasks~\cite{you2016image,yu2017end} to complement the global-level visual cue that are obtained using 2D or 3D CNNs (\eg, ResNet~\cite{resnet}, SlowFast~\cite{slowfast}, etc.).
Inspired by this, we leverage the visual concepts during the fine-tuning stage of individual downstream tasks as an additional language source.
To extract the visual concepts, we employ VinVL~\cite{vinvl}, due to its good performance, which is a detection model providing visual concepts such as objects and attributes.
Given a video, we first sample a frame from individual time intervals of subtitles.
Then, VinVL is applied to the sampled set of frames and provides three visual concept labels over the maximum number of 10 regions for each frame.
These extracted visual concept labels are fed to the text embedding network (\ie, RoBERTa tokenizer followed by a word embedding layer with positional encoding) of HERO model after being attached to subtitles;
the visual concepts of individual regions are separated using [SEP] token.
The examples of extracted visual concepts are illustrated in Figure~\ref{fig:vcept}.

\subsection{Task-aware Ensemble Strategies}
Since the output of a model differs depending on the task, different strategies for ensemble need to be established. For example, we can not simply average the confidence scores in the captioning task to use the simplest form of a model ensemble because the captioning model does not output the confidence score but the predicted caption itself. Therefore, we specialize the ensemble strategies for each task.

\begin{table*}[t]
    \caption{
    Single and Ensemble Model Performance. 
    The baseline column denotes the best score of VALUE baselines shown on the leaderboard.
    CS, M, and RS indicate visual features of Clip-ViT+SlowFast, MIL-NCE, and ResNet+SlowFast, respectively.
    }
    \vspace{+0.12cm}
    \centering
    \begin{adjustbox}{width=0.95\textwidth}
        \begin{tabular}{cccccccccc}
        \toprule
        \# & Macro-Task & Dataset (Task) & V-Concept & AT & ST & Baseline & Ours (Best Single) & Ours (Ensemble) & Metrics \\
        \hline \hline
        M1 & \multirow{4}{*}{Retrieval} & TVR & - & CS & CS & 13.70 & 13.15 & 14.80 (\textcolor{blue}{+12.55\%}) & \multirow{4}{*}{AveR} \\
        M2 & & How2R & $\checkmark$ & - & M & 3.95 & 6.05 & 8.44 (\textcolor{blue}{+39.55\%}) &  \\
        M3 & & YC2R & - & - & M & 56.59 & 54.46 & 57.22 (\textcolor{blue}{+5.07\%}) &  \\
        M4 & & VATEX-EN-R & $\checkmark$ & CS & CS & 49.91 & 76.18 & 73.54 (\textcolor{red}{-3.47\%}) &  \\
         \hline
        M5 & \multirow{4}{*}{QA} & TVQA & $\checkmark$ & - & RS & 74.83 & 75.63 & 77.66 (\textcolor{blue}{+2.68\%}) & \multirow{4}{*}{Acc.} \\
        M6 & & How2QA & $\checkmark$ & - & RS & 74.60 & 77.11 & 79.74 (\textcolor{blue}{+3.41\%}) &  \\
        M7 & & VIOLIN & - & CS & CS & 67.18 & 69.43 & 70.72 (\textcolor{blue}{+1.86\%}) &  \\
        M8 & & VLEP & - & CS & CS & 69.37 & 68.21 & 69.15 (\textcolor{blue}{+1.38\%}) &  \\
         \hline
        M9 & \multirow{3}{*}{Captioning} & TVC & $\checkmark$ & - & CS & 51.04 & 52.30 & 56.17 (\textcolor{blue}{+7.40\%}) & \multirow{3}{*}{CIDEr-D} \\
        M10 & & YC2C & - & - & M & 121.89 & 134.80 & 141.17 (\textcolor{blue}{+4.73\%}) &  \\
        M11 & & VATEX-EN-C & - & CS & CS & 58.09 & 56.36 & 60.17 (\textcolor{blue}{+6.76\%}) & \\
        \bottomrule
        \end{tabular}
    \end{adjustbox}
    \label{tab:single_ensemble_model}
\end{table*}

\begin{table*}[t]
    \caption{Impact of visual feature adaptation. The HERO model that is pre-trained with ResNet+SlowFast (RS) feature is fine-tuned with eight different visual features; ResNet+SlowFast (RS), ResNet+MIL-NCE (RM), Clip-ViT+SlowFast (CS), Clip-ViT+MIL-NCE (CM), ResNet (R), SlowFast (S), Clip-ViT (C), MIL-NCE (M).
    The best and worst numbers are highligted in blue and red respectively.}
    \vspace{+0.12cm}
    \centering
    \begin{adjustbox}{width=1.0\textwidth}
        \begin{tabular}{cccccccccc}
        \toprule
        \multirow{2}{*}{\#} & \multirow{2}{*}{Dataset} & \multicolumn{8}{c}{Feature Transferability (RS $\rightarrow$ Others)} \\
        \cline{3-10}
        &  & RS & RM & CS & CM & R & S & C & M \\
        \hline \hline
        P1 & TVR & \textbf{11.77} & 8.75 (\textcolor{red}{-3.02}) & \textcolor{blue}{12.48} (\textcolor{blue}{+0.71}) & 11.15 & 8.90 & 10.83 & 11.29 & 9.89 \\
        P2 & How2R & \textbf{5.74} & 5.46 & \textcolor{blue}{5.84} (\textcolor{blue}{+0.11}) & 5.64 & 5.07 (\textcolor{red}{-0.67}) & 5.20 & 5.69 & 5.82 \\
        P3 & YC2R & \textbf{49.87} & 48.22 & 48.92 & 49.35 & 47.92 & 47.60 (\textcolor{red}{-2.27}) & 48.52 & \textcolor{blue}{54.46} (\textcolor{blue}{+4.59}) \\
        P4 & VATEX-EN-R & \textbf{61.46} & 46.21 & \textcolor{blue}{66.96} (\textcolor{blue}{+5.50}) & 64.29 & 45.57 (\textcolor{red}{-15.89}) & 58.76 & 63.72 & 51.22 \\
        
        \hline
        P5 & TVQA & \textcolor{blue}{\textbf{74.39}} & 72.10 (\textcolor{red}{-2.29}) & 74.32 (\textcolor{red}{-0.07}) & 72.23 & 73.30 & 73.71 & 73.19 & 73.19 \\
        P6 & How2QA & \textcolor{blue}{\textbf{74.74}} & 72.52 (\textcolor{red}{-2.22}) & 74.03 (\textcolor{red}{-0.71}) & 73.03 & 73.13 & 73.52 & 73.16 & 73.48 \\
        P7 & VIOLIN & \textcolor{blue}{\textbf{67.68}} & 67.26 & 67.67 & 67.60 & 67.00 & 67.37 & 67.67 (\textcolor{red}{-0.01}) & 66.78 (\textcolor{red}{-0.68}) \\
        P8 & VLEP & \textcolor{blue}{\textbf{67.24}} & 65.35 (\textcolor{red}{-1.89}) & 66.10 & 65.76 & 66.07 & 66.51 & 66.58 (\textcolor{red}{-0.66}) & 66.05 \\
        
        \hline
        P9 & TVC & \textbf{49.86} & 46.19 (\textcolor{red}{-3.67}) & 51.24 & 50.62 & 47.22 & 48.96 & \textcolor{blue}{51.34} (\textcolor{blue}{+1.48}) & 48.93 \\
        P10 & YC2C & \textbf{121.80} & 112.20 (\textcolor{red}{-9.60}) & 115.00 & 118.30 & 112.80 & 113.00 & 121.00 & \textcolor{blue}{134.80} (\textcolor{blue}{+13.00}) \\
        P11 & VATEX-EN-C & \textbf{51.39} & 38.68 & \textcolor{blue}{55.42} (\textcolor{blue}{+4.03}) & 50.85 & 37.61 (\textcolor{red}{-13.78}) & 49.70 & 50.93 & 41.36 \\
        \Xhline{2\arrayrulewidth}
        \end{tabular}
    \end{adjustbox}
    \label{tab:feature_transferability}
\end{table*}

\paragraph{Bayesian optimization for retrieval.}
Given $N_r$ retrieval models, we obtain a list of similarity score matrices~\footnote{For Video Corpus Moment Retrieval (VCMR), we apply non-maximum suppression (NMS) with IoU threshold 0.7 to retrieve max. 100 candidates.} $\mathcal{S}_{\text{ret}} = \{\bm{S}_{1}, \bm{S}_{2}, ..., \bm{S}_{i}, ..., \bm{S}_{N_r}\}$ where $\bm{S_i} \in \mathbb{R}^{N_q \times N_g}$ means a similarity matrix between the $N_q$ text queries and $N_g$ candidate videos for the $i^{\text{th}}$ retrieval model.
Then, we find a set of weights $\mathcal{W}_{\text{ret}} = \{w_1, w_2, ..., w_{N_r}\}$ to be used for identifying the best combination of retrieval models.
Finally, the ensembled score matrix $\bm{S}_{\text{ENS}}$ is given by
\begin{equation}
    \bm{S}_{\text{ENS}} = w_1 \bm{S}_{1} + w_2 \bm{S}_{2} + ... + w_{N_r} \bm{S}_{N_r},
\end{equation}
where $\sum_{n}w_n$ = 1. 
To find the optimal values for $\mathcal{W}_{\text{ret}}$, we resort to hyper-parameter tuning by Bayesian optimization\footnote{https://github.com/hyperopt/hyperopt}. To be specific, given that $\mathcal{S}_{\text{ret}}$ is obtained by predictions of $N_r$ models, we set a $N_r$-dimensional uniformly distributed search space. Then, the weights are randomly sampled and fed into Tree-structured Parzen Estimator algorithm~\cite{bergstra2011algorithms} to determine the optimal values for $\mathcal{W}_{\text{ret}}$ based on the objective function.
We leverage the mean recall (\ie, (R@1+R@5+R@10)/3) as the objective function. 
We iterate this optimization process over 300 steps.

\paragraph{Training single-layered FC for QA.}
Given $N_{\text{qa}}$ QA models, our objective is to find the best set of weights $\mathcal{W}_{\text{qa}} = \{w_1, w_2, ..., w_{N_{\text{qa}}}\}$.
Basically, QA task in VALUE is identical to a simple classification that predicts an answer label $c \sim \{c_1, c_2, ..., c_{N_{\text{ans}}}\}$, where $N_{\text{ans}}$ is the number of candidate answers.
In this formulation, the QA models output the confidence scores $\mathcal{S}_{\text{qa}} = \{s_1, s_2, ..., s_i..., s_{N_{\text{qa}}} \}$, where $s_i \in \mathbb{R}^{m \times N_{\text{ans}}}$ is the score matrix representing the confidence scores for each class outputted by $i$-th model given $m$ examples of test set.
Instead of using Bayesian optimization, we use a learning-based approach for QA. We convert the problem to find the optimal $\mathcal{W}_{\text{qa}}$ to learning a single-layered Fully-connected layer (FC) with no bias.
Firstly, we collect all model outputs $\mathcal{S}_{\text{qa}}$ and batchify it to build an input $X \in \mathbb{R}^{B \times N_{\text{qa}} \times N_{\text{ans}}}$, where $B$ is the batch size.
Then we train a linear layer ($\phi$; nn.Linear($N_{\text{qa}}$, 1), bias=False)) to obtain output $\bar{X} = \phi(X), \bar{X} \in \mathbb{R}^{B \times N_{\text{ans}}}$. Lastly, we apply the cross-entropy loss with ground-truth label.
The strength of the learning-based approach is that it converges fast regardless of how large $N_{\text{qa}}$ is used. On the other hand, the disadvantage is evident in that it requires careful hyperparameter tuning tor training.

\begin{table*}[h]
    \caption{Analysis on the effect of visual concepts.}
    \vspace{+0.12cm}
    \centering
    \begin{adjustbox}{width=1.0\textwidth}
        \begin{tabular}{cccccccc}
        \toprule
        \# & { Macro-Task} & { Domain} & { Dataset (Task)} & { Val.} & { $\Delta$ (\%) by dataset} & { Avg. $\Delta$ (\%) by macro-task} & { Avg. $\Delta$ (\%) by domain} \\
        \hline \hline
        G1 & { } & { TV } & { TVR} & { 11.77  $\rightarrow$ 11.69} & \textcolor{red}{ -0.68} & \textcolor{red}{ -0.51} & \textcolor{blue}{ +0.49} \\
        G2 & { } & { HowTo100M} & { How2R} & { 5.73  $\rightarrow$ 5.61} & \textcolor{red}{ -2.09} & \textcolor{red}{ -0.51} & \textcolor{blue}{ +0.54} \\
        G3 & { } & { YouCook2} & { YC2R} & { 49.87  $\rightarrow$ 49.18} & \textcolor{red}{ -1.38} & \textcolor{red}{ -0.51} & \textcolor{red}{ -1.47} \\
        G4 & \multirow{-4}{*}{{ Retrieval}} & { VATEX} & { VATEX-EN-R} & { 61.46  $\rightarrow$ 62.75} & \textcolor{blue}{ +2.10} & \textcolor{red}{ -0.51} & \textcolor{blue}{ +2.10} \\
        \hline
        G5 & { } & { TV } & { TVQA} & { 74.39  $\rightarrow$ 75.63} & \textcolor{blue}{ +1.67} & \textcolor{blue}{ +0.89} & \textcolor{blue}{ +0.49} \\
        G6 & { } & { HowTo100M} & { How2QA} & { 74.74  $\rightarrow$ 77.11} & \textcolor{blue}{ +3.17} & \textcolor{blue}{ +0.89} & \textcolor{blue}{ +0.54} \\
        G7 & { } & { VIOLIN} & { VIOLIN} & { 67.60  $\rightarrow$ 67.37} & \textcolor{red}{ -0.34} & \textcolor{blue}{ +0.89} & \textcolor{red}{ -0.34} \\
        G8 & \multirow{-4}{*}{{ QA}} & { VLEP} & { VLEP} & { 67.24  $\rightarrow$ 66.62} & \textcolor{red}{ -0.92} & \textcolor{blue}{ +0.89} & \textcolor{red}{ -0.92} \\
        \hline
        G9 & { } & { TV } & { TVC} & { 49.86  $\rightarrow$ 50.10} & \textcolor{blue}{ +0.48} & \textcolor{blue}{ +0.34} & \textcolor{blue}{ +0.49} \\
        G10 & { } & { YouCook2} & { YC2C} & { 121.90  $\rightarrow$ 120.00} & \textcolor{red}{ -1.56} & \textcolor{blue}{ +0.34} & \textcolor{red}{ -1.47} \\
        G11 & \multirow{-3}{*}{{ Captioning}} & { VATEX} & { VATEX-EN-C} & { 51.39  $\rightarrow$ 52.47} & \textcolor{blue}{ +2.10} & \textcolor{blue}{ +0.34} & \textcolor{blue}{ +2.10} \\
        \bottomrule
        \end{tabular}
    \end{adjustbox}
    \label{tab:effect_of_vcept}
\end{table*}

\begin{table*}[h]
    \caption{Leaderboard Score Transition at each trials. The asterisk mark ($*$) indicates the prediction results for submission were mistakenly calculated due to internal bug in our code.}
    \vspace{+0.12cm}
    \centering
    \begin{adjustbox}{width=0.70\textwidth}
        \begin{tabular}{ccccccccc}
        \toprule
        \# & Ensemble & V-Concept & AT & ST & Retrieval & QA & Captioning & Meta-Ave \\
        \hline \hline
        T1 & \checkmark &  &  & \checkmark & 34.89 & 67.38 & 35.29$^{*}$ & \textbf{60.63} \\
        T2 & \checkmark &  & \checkmark & \checkmark & 35.29 & 72.21 & 85.95 & \textbf{62.53} \\
        T3 & \checkmark & \checkmark &  & \checkmark & 35.05 & 72.89 & 85.95 & \textbf{62.69} \\
        T4 & \checkmark & \checkmark & \checkmark & \checkmark & 35.02 & 73.01 & 85.95 & \textbf{62.72} \\
        \bottomrule
        \end{tabular}
    \end{adjustbox}
    \label{tab:leaderboard_score_transition}
\end{table*}

\paragraph{Consensus-based ranking for captioning.}
Given $N_c$ captioning models, we generate a set of captions $\mathcal{C}_{N_c}=\{c_1,c_2,...,c_{N_c}\}$ for an input video where individual models generate a caption with a greedy decoding strategy.
Then, we adopt a consensus score $s_n$ for a caption $c_n$ from $n^{\text{th}}$ captioning model, which is defined as an averaged similarity to all the other captions as follows:

\begin{equation}
	s_n=\frac{1}{|\mathcal{C}_{N_c}|-1}\sum_{c' \in {\mathcal{C}_{N_c}} \backslash \{ c_n \}} \text{sim}(c_n, c'),\label{eq:cons_cap}
\end{equation}

where $\text{sim}(c_n,c')$ is the similarity between two captions $c_n$ and $c'$.
We employ five sentence embedding models\footnote{https://github.com/UKPLab/sentence-transformers}\footnote{More specifically, we adopt following 5 models---paraphrase-mpnet-base-v2, stsb-mpnet-base-v2, paraphrase-MiniLM-L3-v2, paraphrase-multilingual-mpnet-base-v2, and paraphrase-TinyBERT-L6-v2.} and the averaged cosine similarities of individual embeddings for captions as the similarity function.
Finally, the caption of the highest consensus score is chosen for our final output.

\section{Experiment}
This section discusses the impact of the single model optimization, fine-tuning with visual concepts on individual tasks, and the task-aware ensemble strategies.

\subsection{Single Model Performance}
As discussed in Section~\ref{sec:single_model_optimization}, we optimize a single model training configuration for 11 tasks individually.
The best configurations (\ie, usage of visual concept, application of AT and ST, and the choice of visual feature) and the corresponding best single model performance are summarized in Table~\ref{tab:single_ensemble_model}.
We can observe the followings.
Firstly, our optimized models outperform the counterparts of VALUE baselines by large margins in most cases, while some of them show slightly lower scores (see M1, M3, M8, and M11).
Secondly, the best combinations of (1) visual feature adaptation during the fine-tuning stage and (2) fine-tuning scheme (\ie, ST or AT$\rightarrow$ST) are highly dependent on both domain and task.
In general, using Clip-ViT+SlowFast (CS) visual feature with the AT$\rightarrow$ST scheme shows outstanding results.
On the other hand, in some cases, a specific type of feature significantly outperforms the others.
For example, we found that fine-tuning on the single task with MIL-NCE feature performs best for the YouCook2 dataset (see M3 and M10);
it is expected because the videos in YouCook2 and HowTo100M (used for pre-training MIL-NCE) share the domain, implying the importance of in-domain pre-training.
Lastly, the task-aware model ensemble further provides performance gain in all tasks except VATEX-EN-R in Retrieval.
In addition, Table~\ref{tab:feature_transferability} shows the impact of visual feature adaptation where it provides the performance gain in Retrieval and Captioning macro-tasks (see P1-4 and P9-11).
The results from the two tables indicate the importance of configuration optimization for individual tasks.

\subsection{Effect of Visual Concepts}
As illustrated in Figure~\ref{fig:vcept} and Section~\ref{sec:transfer_learning_vcept}, we extract the visual concepts from raw frames and leverage them with the corresponding subtitles as auxiliary information. Since we have a lot of missing videos that failed to download, a limited portion of videos in each dataset domain (\ie, YouCook2~(89\%), VIOLIN~(15\%), VATEX~(76\%), HowTo100M~(87\%), and TV ~(99\%)) is used for the extraction. 
For training samples from the missing videos, we use the subtitle only without the visual concepts. Table~\ref{tab:effect_of_vcept} analyzes the effect of using the visual concepts during the fine-tuning. We observe the performance enhancement on five out of eleven tasks (see G4-6, G9, and G11) compared to without employing the visual concepts. 
We found that fine-tuning with the visual concepts is especially effective for VATEX videos (both retrieval and captioning; G4 and G11) as well as for QA macro-tasks (G5-8).
On the other hand, fine-tuning with visual concepts did not improve the performance on the retrieval task. Although the effectiveness of visual concepts depends on many experimental variables (\eg, domains, tasks, \# of visual concepts, etc.), it helps to increase the variance between models, which is known to be an essential factor for the compelling model ensemble.

\subsection{Model Ensemble}
Our submission scores in Table~\ref{tab:leaderboard_score_transition} are obtained by model ensemble. Given fine-tuned models trained with various training configurations, we choose the top $K$ models sorted by the validation scores for the ensemble.
Note that we vary the hyperparameter $K$ according to the macro-task because we found a trade-off between computational cost and the performance for the ensemble. We set $K$ to 8, 16, and 32 for captioning, QA, and retrieval macro-tasks, respectively. Table~\ref{tab:single_ensemble_model} shows the validation score of our ensemble model. In most cases, we achieve significant score improvement with a model ensemble by large margins (max. +39.55\% and avg. +7.45\%). The retrieval task benefits the most from the ensemble, followed by captioning and QA.

\section{Conclusion}
We described our winning strategies for the VALUE challenge 2021. To solve the task, we propose three main key ingredients: 1) single model optimization, 2) transfer learning with visual concepts, and 3) task-aware ensemble. Through the proposed strategies, the score is improved step by step as shown through extensive experiments in this report. Our final submission is an ensemble of the $K$ number of best performing single models on the validation set, that are trained with various training configurations. Based on our approach, we achieve 1st place on the VALUE and QA phases for the competition.

{\small
\bibliographystyle{ieee_fullname}
\bibliography{egbib}
}

\end{document}